\documentclass{article}

\usepackage{graphicx}
\usepackage{amsmath}
\usepackage{booktabs}
\usepackage{subcaption} 
\usepackage{caption} 
\usepackage{url}
\usepackage{natbib}
\newcommand{\minus}{\scalebox{0.75}[1.0]{$-$}}
\usepackage[autostyle]{csquotes}
\usepackage[british]{babel}

\begin{document}
\title{A comparison of classical and variational autoencoders for anomaly detection\\
{\small Technical Report IDSIA-2020-10.1}}
%\small{Technical Report IDSIA-2020-10.1}
\author{Fabrizio Patuzzo 
\\[6pt]
IDSIA / USI-SUPSI
\\[2pt]
Manno, Switzerland
%fabrizio.patuzzo@idsia.ch
%\href{fabrizio.patuzzo@idsia.ch}
}
\date{September 2020}

\bibliographystyle{abbrvnat}
\maketitle

\begin{abstract}
This paper analyzes and compares a classical and a variational autoencoder in the context of anomaly detection. To better understand their architecture and functioning, describe their properties and compare their performance, it explores how they address a simple problem: reconstructing a line with a slope. %, by analyzing how they solve a simple reconstruction problem. 
\end{abstract}

\section*{Introduction}
%The aim of this short paper is to compare the performance of a classical and a variational autoencoder. 
An autoencoder is a neural network that receives an input (usually an image), compresses it into a short code (the \textquote{bottleneck}) and outputs a reconstruction of the image. They are used to allow computers to produce works of art (by modifying the values in the bottleneck, the network creates new, unseen pictures), and to detect anomalies: an autoencoder can be trained to reproduce properly only clean images, but not defective ones, making it possible to detect the defect. There exist two types of autoencoders: classical and variational. Variational autoencoders have been particularly successful to generate art.%\footnote{See, for example, \citet{Lindbo2015}}. %\footnote{While variational autoencoders have been successfully used to generate art, their use in anomaly detection is limited.}. 
\\[6pt]
The aim of this short paper is to better understand how classical and variational autoencoders work in the anomaly detection domain, and to compare their performance. To do this, we will look at how they tackle a simple problem: reconstructing a line with a slope. %\footnote{Other comparisons between a classical and variational autoencoder include \cite{AnCho2015}.}. 
%anomaly detection applications, see \cite{Chalapathy2019}.}.
%Moreover, we would like to take a look inside, to better understand how they work - just like a mechanic might look inside a car's engine to debug problems or enhance its performance. 
%open an autoencoder used for anomaly detection and have a look inside. The hope is to better understand how it works, how to choose an architecture and hyperparameters suitable for a given task, and why sometimes the training fails - just like a mechanic looks inside a car's engine to debug problems or enhance its performance.  
\section{Training Set}
\begin{figure}[ht!]
\centering
\includegraphics[scale=0.7]{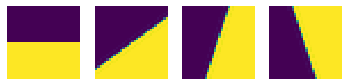}
\caption{A few samples from the training set}
\label{fig:training_set1}
\end{figure}

The training set is composed of 1000 black and white, 32x32 pixel images, representing a straight line with a given slope that divides the image into white and black halves. Figure~\ref{fig:training_set1} shows a few images from the training set (plotted using the colormap \textquote{viridis}). %\footnote{Viridis maps white to yellow, black to violet and gray to green.}. 

\section{Classical Autoencoder}

\begin{figure}[ht!]
\centering
\includegraphics[scale=0.7]{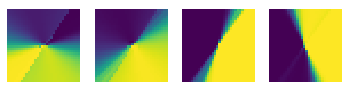}
\caption{Results using A1}
\label{fig:results_A1}
\end{figure}

The classical autoencoder (CA) is a Multi Layer Perceptron (MLP) with the following architecture:
\begin{table}[ht!]
\begin{center}
\begin{tabular}{ l c c c c c }
\toprule
LAYER & type & neurons & activation \\ 
\midrule
0 & Input & 32x32 & \\  
1 & Dense & 2 & relu\\ 
2 (bottleneck) & Dense & 1 & \\ 
3 & Dense & 2 & relu\\  
4 & Dense & 32x32 & sigmoid\\ 
 \bottomrule
\end{tabular}
\end{center}
\vspace{-4mm}
\end{table}
\\[6pt]
The input is a 32x32 px black and white image, and all layers are Dense. L1 and L3 contain two neurons and are followed by a \emph{relu} activation function, defined as $max(0, x)$. L2, the bottleneck, contains only one neuron. The final layer is followed by a \emph{sigmoid} function: $\frac{e^x}{e^x+1}$.
\\[6pt]
We trained the CA to recompose the input images in the output. Figure~\ref{fig:results_A1} shows the results. The reconstructions are not precise: there is some blur along the dividing lines, and some sort of strange kaleidoscopic effects in the first two images. 

\begin{figure}[ht!]
\centering
\includegraphics[scale=0.5]{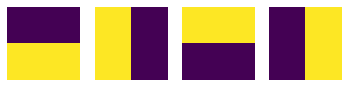}
\caption{Simplified training set}
\label{fig:dataset2}
\end{figure}

To try to pinpoint the cause of this blur, we have created an even simpler dataset composed of four 2x2 images, where the slope can either be horizontal or vertical, shown in Figure~\ref{fig:dataset2}.
\\[6pt]
We trained the CA again (adapting its input and output dimensions) and, after several failed attempts (more on this later), it finally succeeded in reconstructing the inputs: \textquote{0011}, \textquote{1010}, \textquote{1100} and \textquote{0101}\footnote{the four digits correspond to the pixel values in the upper left, upper right, bottom left and bottom right corners.}. 
\\[6pt]
Let us see which weights the network has learned\footnote{To simplify the discussion, we will assume that all biases in L1, L2 and L3 equal zero.}.  
\begin{enumerate}
\item Figure~\ref{fig:output1} shows the weights leading to the first output neuron.
\begin{figure}[ht!]
\centering
\fbox{
\includegraphics[scale=0.5]{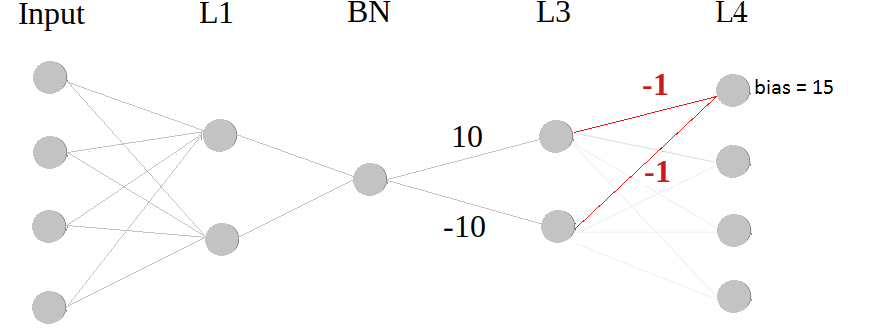}
}
\caption{weights leading to output neuron 1}
\label{fig:output1}
\end{figure}
\\[6pt]
The output of the first neuron in L4 as a function of the bottleneck is:  
\begin{equation}
y=\sigma(relu(10 x)\cdot(\minus 1)+relu(\minus 10 x)\cdot(\minus 1)+b). 
\end{equation}
where $\sigma$ is the sigmoid, $x$ is the bottleneck value and $b$ the bias. 
\\[6pt]
Let us approximate the sigmoid with a Heaviside function, which returns $0$ if $x<0$, and $1$ if $x\geq0$. 
\\[6pt]
Then, when the bottleneck equals $\minus 2$, the output is $H(\minus 5)=0$; when it equals $0$ the output is $H(15)=1$; and a bottleneck of $2$ generates an output of $H(\minus 5)=0$.
\\[6pt]
We can express this relation in tabular form.%\footnote{To express the relation in tabular form, we approximate the sigmoid with a Heaviside function (which returns $0$ if $x<0$, and $1$ if $x\geq0$).}:
\begin{table}[ht!]
\begin{center}
\begin{tabular}{l c c c c c c c}
\toprule
BN & \minus $\infty$ & & \minus 1.5 & & 1.5 & & $+\infty$ \\ 
\midrule
output 1 & & 0 & & 1 & & 0 &\\  
 \bottomrule
\end{tabular}
\end{center}
\vspace{-4mm}
\end{table}
\\[6pt]
The thresholds, $\minus 1.5$ and $1.5$,  can be obtained by computing $t_1=\minus \frac{b}{10}$ and $t_2=\frac{b}{10}$.  
\\[6pt]
%A bottleneck of $-2$ generates an output of $-5$. A bottleneck of $0$, an output of $15$. And a bottleneck of $2$, an output of $-5$. 
\item Figure~\ref{fig:output2} illustrates the weights leading to the second output neuron.
\begin{figure}[ht!]
\centering
\fbox{
\includegraphics[scale=0.5]{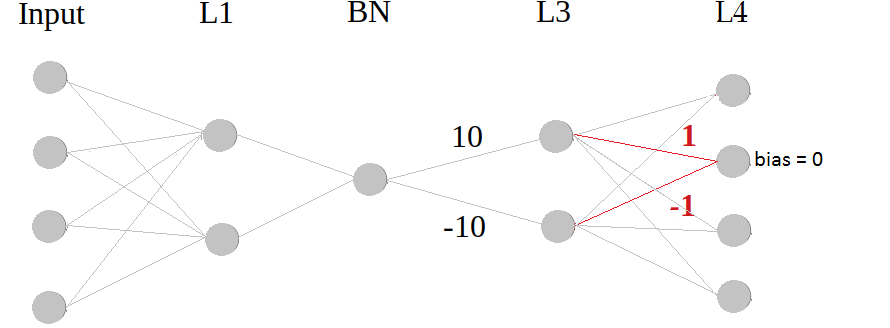}
}
\caption{weights leading to output neuron 2}
\label{fig:output2}
\end{figure}

This time, any strictly negative bottleneck will output a $0$, and any positive bottleneck will output a $1$. 
\\[6pt]
The relation between the bottleneck and output 2 is:
\begin{table}[ht!]
\begin{center}
\begin{tabular}{ l c c c c c c c c c}
\toprule
BN & \minus $\infty$ & & & & 0 & & & & $+\infty$ \\ 
\midrule
output 2 & & & 0 & & & & 1 & &\\  
 \bottomrule
\end{tabular}
\end{center}
\vspace{-4mm}
\end{table}

\item In Figure~\ref{fig:output3}, one can see the connections from the bottleneck to output neuron 3.
\begin{figure}[ht!]
\centering
\fbox{
\includegraphics[scale=0.5]{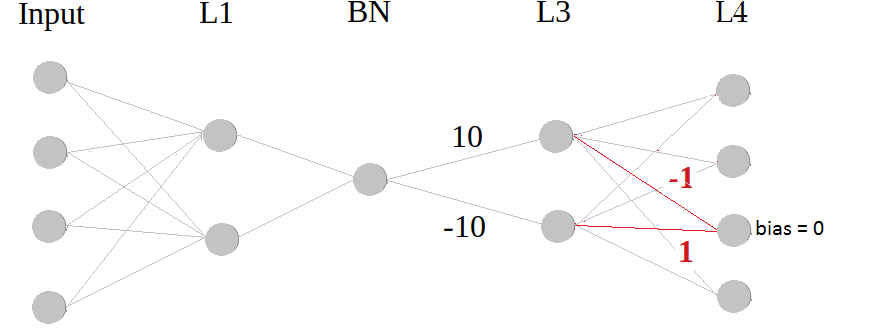}
}
\caption{weights leading to output neuron 3}
\label{fig:output3}
\end{figure}
\\[6pt]
This is exactly the reversed of output 2. Notice that in all the inputs, the second and the third digit are complementary.
\\[6pt]
The relation between BN and output 3 is:
\begin{table}[ht!]
\begin{center}
\begin{tabular}{ l c c c c c c c c c}
\toprule
BN & \minus $\infty$ & & & & 0 & & & & $+\infty$ \\ 
\midrule
output 3 & & & 1 & & & & 0 & &\\  
 \bottomrule
\end{tabular}
\end{center}
\vspace{-4mm}
\end{table}
\\[6pt]
\item Figure~\ref{fig:output4} shows the weights that lead from the bottleneck to output neuron 4.
\begin{figure}[ht!]
\centering
\fbox{
\includegraphics[scale=0.5]{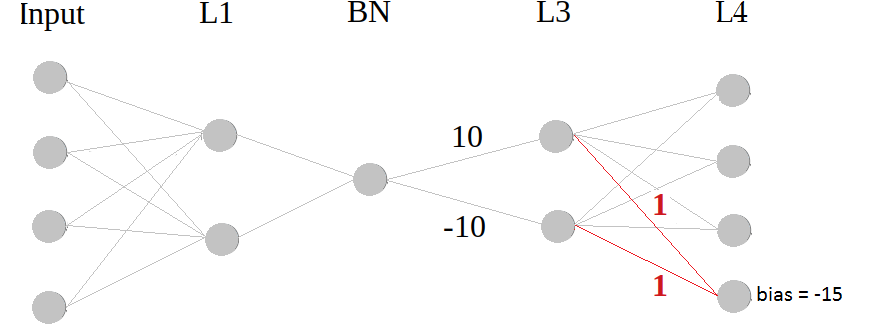}
}
\caption{connections leading to output neuron 4}
\label{fig:output4}
\end{figure}
\\[6pt]
This time, output 4 is the opposite of output 1 and the relation between BN and output 4 is:
\begin{table}[ht!] 
\begin{center}
\begin{tabular}{ l c c c c c c c}
\toprule
BN & \minus $\infty$ & & \minus 1.5 & & 1.5 & & $+\infty$ \\ 
\midrule
output 4 & & 1 & & 0 & & 1 &\\  
 \bottomrule
\end{tabular}
\end{center}
\vspace{-4mm}
\end{table}
\end{enumerate}

\begin{figure}[ht!]
\centering
\includegraphics[scale=0.5]{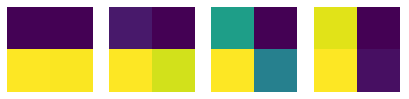}
\caption{outputs as BN crosses the critical value $1.5$}
\label{fig:bottleneck}
\end{figure}

\begin{figure}[ht!]
\centering
\includegraphics[scale=0.45]{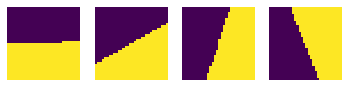}
\caption{outputs after the postprocessing}
\label{fig:A1_post}
\end{figure}

If we increase BN evenly from $\minus \infty$ to $+\infty$, the network will output a copy of the inputs: \textquote{0011}, \textquote{1010}, \textquote{1100} and \textquote{0101}. The output will change when BN crosses the thresholds $\minus 1.5$, $0$ and $1.5$.  %and $-1.5$, one obtains the first image: 0011. With a bottleneck value between $-1.5$ and $0$, one obtains the second: 1010. With a value between $0$ and $+1.5$, the third (1100), and with a value between $+1.5$ and $+\infty$, the fourth: 0101. 
\\[6pt]
But let us remember, we have approximated the sigmoid with a Heaviside function. The only difference if we use a true sigmoid is that as the bottleneck approaches a threshold, some output neurons will become temporarily gray, as shown in Figure~\ref{fig:bottleneck}.  
\\[6pt]
Hence, some \textquote{blur} can occur when BN is very close to a threshold. To remove the blur, we can replace the sigmoid with a Heaviside function. %However, since we needed the sigmoid to train the network, we can compute the inverse sigmoid of the output, and then the Heaviside. 
%In this way, the blur disappears. 
\\[6pt]
%Is this the same reason why the outputs were blurry applying CA to the initial dataset? And can we remove it in the same way? Yes! 
Back to our initial dataset. The logic of the network is exactly the same, only there are more output neurons, each one with its own bias $b$ that will determine at which thresholds $t_1$ and $t_2$ the neuron value changes. As we increase the bottleneck value, some neurons will change value and become temporarily gray, causing the blur in the figure.  
\\[6pt]
Let us try to apply the same trick to remove the blur. 
\\[12pt]
By replacing the sigmoid with a Heaviside function after the training, the blur disappears, as well as the strange kaleidoscopic effects, as one can see in Figure~\ref{fig:A1_post}. 

\section{Variational Autoencoder}
\begin{figure}[ht!]
\centering
\fbox{
\includegraphics[scale=0.4]{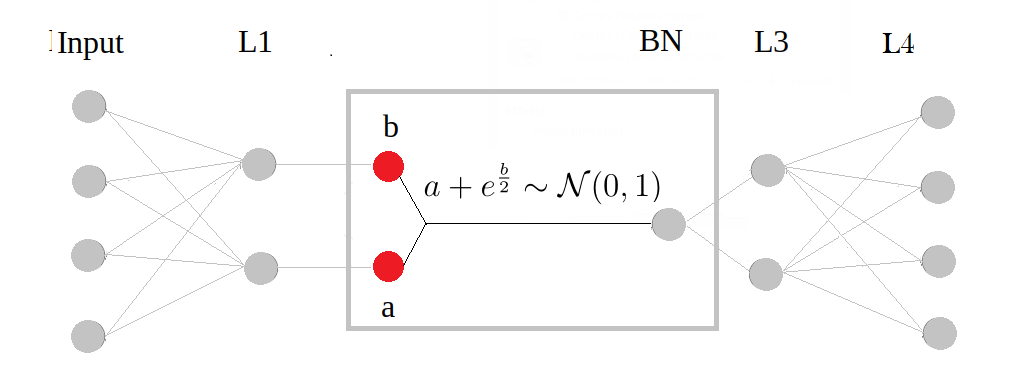}
}
\caption{The variational architecture}
\label{fig:variational}
\end{figure}

Let us see now how a variational autoencoder (VA) would perform. 
\\[6pt]
The architecture of a VA is similar to the architecture of a CA, but with the little modification shown in Figure~\ref{fig:variational}\footnote{Our implementation of the variational autoencoder is inspired by \cite{Chollet2019}.}.

A new layer is added just before the bottleneck, composed of two neurons, $a$ and $b$. The bottleneck value is created by picking it from a random normal distribution with mean $a$ and variance $e^b$\footnote{Notice that in \citet{Chollet2016} the implementation is slightly different, and $e^b$ represents the standard deviation rather than the variance.}. 
\\[6pt]
Not only the architecture, but also the \emph{loss} function used to train the network is different. In a CA it is just the mean squared error (\emph{mse}), but in a VA it includes a new term\footnote{The mathematical derivation of the new term, obtained computing the Kullback-Leibler divergence between two normal distributions, is explained in \citet{Kingma2014}, the paper that set the mathematical foundations of the variational architecture.}: 
\begin{equation}
loss = mse - \frac{1}{2n} \cdot \sum_{1}^{n} (1 + b - a^2 - e^b)
\end{equation}
where $n$ is the number of training samples. 
\\[6pt]
The two changes in the architecture and in the \emph{loss} have an impact on the characteristics of the autoencoder:  
\begin{enumerate}
\item Every input image now produces a bottleneck value picked from a normal distribution with $\mu=a$ and $\sigma^2=e^b$. 
\item The new \emph{loss} ensures that the bottleneck values generated by the training set will follow a normal distribution with $\mu = 0$ and $\sigma^2=1$, and, as a consequence, they will remain within the range [\minus 3, 3] with a probability of 0.97. 
\end{enumerate}

\begin{figure}[ht!]
\centering
\begin{subfigure}{.5\textwidth}
  \centering
  \includegraphics[width=.8\linewidth]{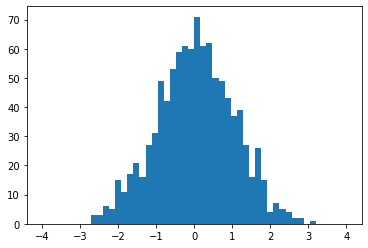}
  \caption{The bottleneck means}
  \label{fig:sub1}
\end{subfigure}%
\begin{subfigure}{.5\textwidth}
  \centering
  \includegraphics[width=.8\linewidth]{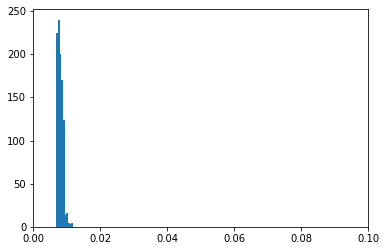}
  \caption{The bottleneck standard deviations}
  \label{fig:sub2}
\end{subfigure}
\caption{}
\label{fig:test}
\end{figure}

After training VA with our dataset, we could verify that the bottleneck values approach a standard normal distribution, as shown in Figure~\ref{fig:sub1}. In Figure~\ref{fig:sub2} one can see the standard deviations ($\sigma$). They are all around 0.01. 

\begin{figure}[ht!]
\centering
\includegraphics[scale=0.6]{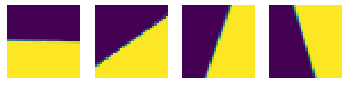}
\caption{Outputs using VA}
\label{fig:results_A2}
\end{figure}

From the bottleneck on, the CA and VA are identical. Hence, we might expect that a VA will find the same kind of weights as a CA. However, since the bottleneck values are packed together in the interval [\minus 3,3], we might expect that they will be close to the thresholds, and thus cause more blur. Plus, they are generated randomly, so our first guess is that the outputs can only be more blurry.  
\\[6pt]
Let us try, and apply a VA to reconstructing the four sample images of the training set. The results are shown in Figure~\ref{fig:results_A2}. They are nearly perfect: no blurring, and no strange kaleidoscopic effects. 
\\[6pt]

Our first guess was wrong. Why is the output not more blurry? And why is it even better?  
\\[6pt]
The fact that the bottlenecks are close together can be easily compensated by finding big weights in L3 and L4, which the network did. And the standard deviations are small enough to ensure that a bottleneck will not cross a threshold. 
%compared to the distance between thresholds. \footnote{It found weights around 10 or -10 in both L3 and L4, which will magnify the distance between thresholds by a factor of 100.}. 
%\\[6pt]
%The weights in L3 and L4 are all around 10 or -10. This means that if two bottlenecks are separated by a gap of 0.06, this gap will be magnified a 100 times in the output, producing outputs that are either white, or black. 
\\[6pt]
But why did the VA perform even better? Because it found a set of optimal weights which the CA did not find, as it converged to a suboptimal solution. %It was unable to find high weights in L4.   
%\\[12pt]
%While we were mainly interested in applying A2 to our initial dataset, we are still curious to see how it would perform on our simplified dataset of 2x2 images. The answer was slightly surprising: its performance is catastrophic! 
%\\[6pt]
%Figure~\ref{fig:results2_A2} shows the reconstructions of the input images of the simplified dataset by A2. 
%%\begin{figure}[h!]
%\centering
%\includegraphics[scale=0.5]{img_var5.png}
%\caption{Results using A2 on the simplified dataset}
%\label{fig:results2_A2}
%\end{figure}
%\\[6pt]
%Not only A2 failed to properly reconstruct the inputs, but the outputs would be different at every run of the autoencoder.
%\\[6pt]
%The reason is that while using the initial dataset (which contained 1000 images) the standard deviations equalled roughly 0.01, using the reduced dataset of four images, the standard deviations increased to values of roughly 0.5 - leading to largely unpredictable outputs. 
%\\[6pt]
%This happened because a variational autoencoder is designed to ensure that the sum of the squares of the standard deviations over the training dataset equals roughly 1. With only four input images, the standard deviations become way too big. 
%\\[6pt]
%This suggests that a variational autoencoder, while it might perform better than a classical one with a large training set, it will not perform well with a small training set.

\section{Converging to a suboptimal solution}
\begin{figure}[ht!]
\centering
\fbox{
\includegraphics[scale=0.5]{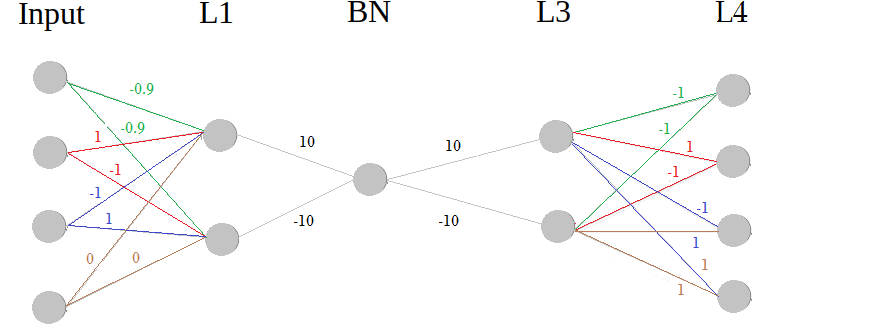}
}
\caption{optimal solution}
\label{fig:opt}
\end{figure}

Even when trained on the simplified dataset of four 2x2 images, the CA did not always find an optimal solution. Figure~\ref{fig:opt} shows what an optimal solution might look like.  
\\[6pt]
This set of weights entirely solves the problem, and the four inputs \textquote{0011}, \textquote{1010}, \textquote{1100} and \textquote{0101} will output copies of themselves, and generate bottleneck values of $\minus 10$, $\minus 1$, $1$ and $10$. 
\\[6pt]
But despite the fact that optimal weights exist, when we train the CA it rarely finds them. Let us see some of the solutions it converges to. 
\begin{enumerate}
\item With an observed frequency of $\frac{1}{8}$, the CA did find a set of optimal weights.
%\begin{figure}[h!]
%\centering
%\includegraphics[scale=0.5]{inputs.png}
%\caption{Optimal solution}
%\label{fig:opt}
%\end{figure}
\item About $\frac{1}{3}$ of the times, it failed to reconstruct two images, producing an \textquote{average} of the two, as shown in Figure~\ref{fig:so1}. 
\begin{figure}[ht!]
\centering
\includegraphics[scale=0.5]{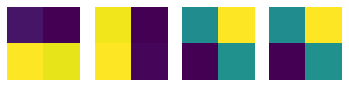}
\caption{suboptimal solution 1}
\label{fig:so1}
\end{figure}

What happened? After weight initialization, two inputs happened to generate negative values in all L1 neurons (which the relu converted to zeros). As a consequence, they produced the exact same output, and no weight update could improve the situation.\footnote{The probability that an input will generate a negative number in at least one neuron in L1 after a random weight initialization is $\frac{1}{4}$, and the probability that this happens to at least two neurons is $1-(3/4)^4-4\cdot(3/4)^3\cdot1/4\approx0.26$}  
%The probability that this happens to at least two input images in the training set is $1-((3/4)^4+4\cdot(3/4)^3\cdot1/4\approx0.26$%\footnote{More generally, with a dataset size of k and n neurons in L1, the probability that this happens is $1-(\binom{0}{k}\cdot(1-p)^k+\binom{1}{k}\cdot(1-p)^{(k-1)}\cdot p$, where $p=(1/2)^n$.}.%15/16)^6\approx0.32$. 
\item Occasionally, this happened with two pairs of images, as illustrated in Figure~\ref{fig:so2}. 
\begin{figure}[ht!]
\centering
\includegraphics[scale=0.7]{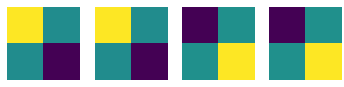}
\caption{suboptimal solution 2}
\label{fig:so2}
\end{figure}
%\\[6pt]
%In this (rare) case, two pairs of two inputs produced the same values in the bottleneck. 
\item About $\frac{1}{3}$ of the times, all the outputs were completely gray, as shown in Figure~\ref{fig:so3}. 
\begin{figure}[ht!]
\centering
\includegraphics[scale=0.5]{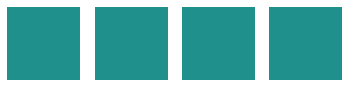}
\caption{suboptimal solution 3}
\label{fig:so3}
\end{figure}

This can happen if all weights in L1 are negative. But there can be other reasons. For example, it can also happen if both weights in L2 are negative and both weights in L3 are positive (or viceversa)\footnote{This last scenario happens $\frac{1}{8}$ of the times right after weight initialization}. %This scenario will happen with a frequency of $1/8=0.125$.%\footnote{More generally, it will happen $2\cdot (1/2)^m \cdot(1/2)^n$, where m and n are the number of neurons in L1 and L3}.
%\item But it is sufficient that one of the values out of L3 be always negative for something bad to happen. In fact, in this case the network will not have enough neurons to find the optimal solution, and might converge to a solution like the one shown in Figure~\ref{fig:three}. This happens with a  frequency of 1/4.  
%\begin{figure}[h!]
%\centering
%\includegraphics[scale=0.5]{three.png}
%\caption{suboptimal solution 4}
%\label{fig:three}
%\end{figure}
\end{enumerate}

These are some of the problems that can happen. In all these cases, two inputs generated the same values after some layer in the network.  
\\[6pt]
One way to reduce the number of times this happens is to increase the number of neurons in L1, L2 and L3 (but too many neurons will cause other difficulties). 
%\footnote{In this case, the probability that one of the above problems arises is reduced to 0.006.}.
%Adding too many neurons in L1, however, might have other side-effects. 
%\\[6pt]
%Let us try to increase the number of inner neurons also to solve our initial %%problem. With 7 neurons in L1 and L3, the probability that the above %mentioned problems happen drops below 0.001, and A1 quickly finds the optimal %solution, as shown in Figure~\ref{fig:A1b}. 
%%\begin{figure}[h!]
%%\centering
%\includegraphics[scale=0.5]{A1b.png}
%\caption{A1 finds the optimal solution}
%\label{fig:A1b}
%\end{figure}
\\[6pt]
Another way, possibly, is to use a VA. For, due to its properties, a VA will never produce exactly the same bottleneck values, preventing the training from stalling. Our guess is that this is the reason why the VA output sharper images.

\section{Conclusions}
We have explored how a classical and a variational autoencoder deal with a simple problem, trying to describe their behaviour and characteristics.  
\\[6pt]
The CA often converged to suboptimal solutions. It reconstructed the images with some blur and some kaleidoscopic effects, even though, in this case, we could remedy by replacing the sigmoid with a Heaviside function. 
\\[6pt]
In contrast, the VA found much better solutions, and we suggested that this was possibly due to the random generation of bottlenecks.
\\[6pt]
If one gives an input to the VA that contains a defect or does not represent a line, it will still output a line, and the reconstruction error will allow to detect the anomaly. 
\\[6pt]
In conclusion, our initial doubts about the ability of a VA to perform better than a CA in anomaly detection were unfounded. %\footnote{\cite{AnCho2015} also proposed a comparison between a CA and a VA, but they used a modified version of a VA.}. 

%performed better than the classical one, even without a postprocessing step.
%  However, we remarked that the variational autoencoder seems to perform poorly when the training set is too small.
%  The classical autoencoder often converged to suboptimal solutions. We addressed this issue by slightly increasing the number of inner neurons. The variational architecture might help avoid some of these problems.
%  Also on the mnist dataset, the variational autoencoder achieved better and sharper results.
%\end{itemize}
%\begin{enumerate}
 %   \item A classical autoencoder can produce blurry edges in the output. In our case, we could solve this issue with a postprocessing step. 
  %  \item The variational autoencoder produced better results than the classical one. 
   % \item Variational autoencoders will not produce good results if the training set is small. 
%    \item It might be necessary to train an autoencoder many times, before it converges to the optimal solution.   
 %   \item Even if there is a minimal architecture that solves the problem, it can be useful to add some neurons in the inner layers to reduce the  number of times the training converges to a suboptimal solution. 
%\end{enumerate}
\section*{Acknowledgments}
We would like to thank our supervisor Alessandro Giusti, and our colleagues Jamal Saeedi and Gabriele Abbate for their useful comments and feedback.

%\newpage

% Literature
%2015: An, J., & Cho, S. Variational autoencoder based anomaly detection using reconstruction probability. 
%Tutorial on variational autoencoders
%Harris Partaourides, Asymmetric deep generative models
%2017: Improved Variational Inference with Inverse Autoregressive Flow
%Generating Faces with Torch
%Encoder: q_phi(z/x) (approximate posterior)
%Decoder: p_theta(x/z) or generative model
%posterior: p_theta(z/x)
%codice ispirato da: %https://github.com/keras-team/keras/blob/master/examples/variational_autoencoder.py
%altro codice: https://github.com/piyush-kgp/VAE-MNIST-Keras/blob/master/vae.py
%\bibliographystyle{apalike}

\bibliography{biblio}

\end{document}